\documentclass[letterpaper, 10 pt, journal, twoside]{ieeetran}

\IEEEoverridecommandlockouts                              %

\def\outputstyle{arxiv}

\usepackage{amsmath} %
\usepackage{amssymb}  %

\usepackage{booktabs}
\usepackage{multirow}
\usepackage[mediumspace,mediumqspace,squaren]{SIunits}
\usepackage{listings}
\makeatletter
\def\lst@makecaption{%
  \def\@captype{table}%
  \@makecaption
}
\makeatother
\usepackage{tikz}
\usetikzlibrary{arrows.meta}
\usetikzlibrary{calc}
\usetikzlibrary{shadows}
\usepackage{xspace}
\usepackage{hyperref}
\usepackage[nameinlink,capitalise]{cleveref}

\xspaceremoveexception{-}
\title{
ros2\_tracing: Multipurpose Low-Overhead Framework for Real-Time Tracing of ROS 2
}

\author{Christophe Bédard$^{1}$, Ingo Lütkebohle$^{2}$, Michel Dagenais$^{1}$%
\thanks{Parts of this work were supported through funding from the European Union's Horizon 2020 research and innovation programme under grant agreement No 780785.
Further, financial support of Ericsson, NSERC, and Prompt for C. Bédard and M. Dagenais is gratefully acknowledged.}%
\thanks{$^{1}$C. Bédard and M. Dagenais are with the Department of Computer Engineering and Software Engineering, Polytechnique Montréal, Montreal, Quebec H3T 1J4, Canada,
        {\tt\scriptsize \{christophe.bedard,michel.dagenais\}@polymtl.ca}}%
\thanks{$^{2}$I. Lütkebohle is with Corporate Research at Robert Bosch GmbH, 71272 Renningen, Germany,
        {\tt\scriptsize ingo.luetkebohle@de.bosch.com}}%
}

\IEEEpubid{%
\scriptsize%
\setlength{\fboxrule}{0pt}%
\fbox{\parbox{0.8\textwidth}{%
\begin{center}%
\vspace{27pt}%
\copyright 2022 IEEE. Personal use of this material is permitted. Permission from IEEE must be obtained for all other uses, in any current or future media, including reprinting/republishing this material for advertising or promotional purposes, creating new collective works, for resale or redistribution to servers or lists, or reuse of any copyrighted component of this work in other works.
\end{center}%
}}}

\begin{document}

\def\ieee{ieee}
\def\arxiv{arxiv}

\newcommand{\rostwotracing}{\texttt{ros2\_tracing}\xspace}
\newcommand{\tracetoolsanalysis}{\texttt{tracetools\_analysis}\xspace}
\newcommand{\ROSone}{ROS~1\xspace}
\newcommand{\ROStwo}{ROS~2\xspace}
\newcommand{\performancetest}{performance\_test\xspace}

\ifx\outputstyle\ieee
    \newcommand{\vspacefiguretikz}{\vspace{-0.3cm}}
    \newcommand{\vspacefigureimage}{\vspace{-0.25cm}}
    \newcommand{\vspacetablefootnotes}{\vspace{-0.25cm}}
    \newcommand{\repourl}{\href{https://gitlab.com/ros-tracing/ros2_tracing}{https://gitlab.com/ros-tracing/ros2\_tracing}}
\else
    \newcommand{\vspacefiguretikz}{\vspace{-0.3cm}}
    \newcommand{\vspacefigureimage}{\vspace{-0.32cm}}
    \newcommand{\vspacetablefootnotes}{\vspace{-0.25cm}}
    \newcommand{\repourl}{\href{https://github.com/ros2/ros2_tracing}{https://github.com/ros2/ros2\_tracing}}
\fi

\newcommand\footnotetextspace{\hspace{2pt}}

\newcommand*\circled[1]{\raisebox{.5pt}{\textcircled{\raisebox{-.9pt} {#1}}}}
 
\maketitle

\begin{abstract}
Testing and debugging have become major obstacles for robot software development, because of high system complexity and dynamic environments.
Standard, middleware-based data recording does not provide sufficient information on internal computation and performance bottlenecks.
Other existing methods also target very specific problems and thus cannot be used for multipurpose analysis.
Moreover, they are not suitable for real-time applications.
In this paper, we present \rostwotracing, a collection of flexible tracing tools and multipurpose instrumentation for \ROStwo.
It allows collecting runtime execution information on real-time distributed systems, using the low-overhead LTTng tracer.
Tools also integrate tracing into the invaluable \ROStwo orchestration system and other usability tools.
A message latency experiment shows that the end-to-end message latency overhead, when enabling all \ROStwo instrumentation, is on average 0.0033~ms, which we believe is suitable for production real-time systems.
\ROStwo execution information obtained using \rostwotracing can be combined with trace data from the operating system, enabling a wider range of precise analyses, that help understand an application execution, to find the cause of performance bottlenecks and other issues.
The source code is available at: \repourl.
 \end{abstract}

\begin{IEEEkeywords}
Software tools for robot programming, distributed robot systems, Robot Operating System (ROS), performance analysis, tracing. \end{IEEEkeywords}

\section{Introduction}\label{sec:introduction}

\ifx\outputstyle\ieee
    \IEEEPARstart{A}{s}
\else
    As
\fi
modern robots have become more versatile, e.g., in tackling unstructured environments or collaborative work, their software has become correspondingly more complex.
Distributed, asynchronous compute graphs based on frameworks like ROS~\ifx\outputstyle\ieee\cite{quigley2009ros,ros2docs}\else\cite{quigley2009ros,macenski2022ros2}\fi\xspace are now the dominant approach for integrated systems, even for space exploration~\cite{perron2021viper}.
Correspondingly, testing and debugging is now typically conducted by collecting data from a running system for later analysis, through tools like rosbag or textual logging~\cite{quigley2015debugging,afzaal2020challenges}.

However, there are well-known drawbacks: rosbag and similar middleware-based tools can only record data that is available as messages.
Aside from the effort involved, there is also a significant resource cost in both CPU and memory usage.
It is also well known that perturbing a system through extensive monitoring is to be avoided~\cite{gregg2020systems}.
Therefore, messages simply cannot practically deliver the stacktrace-level of detail and the detailed execution context information that we have come to expect from classical debuggers.
Logging also cannot fill this gap, because of its unstructured output and lack of support for binary data.
As a result, the debugging experience in robotics is greatly impoverished.

In contrast, tracing has been developed to provide structured, flexible, on-demand data capture across multiple applications and the kernel, to enable detailed analysis when needed.
Conceptually, it can be considered an evolution of logging with support for binary data and well-defined data structures.
Common frameworks provide support for easy and low-overhead capture of contextual data, such as process information or accurate, in-process timestamps, as well as aggregation of data across hosts and tooling for analysis~\cite{gebai2018survey,desnoyers2006lttng,poirier2010accurate,gelle2021combining}.

However, two challenges need to be solved to truly improve the testing and debugging situation.
First, tracing has so far been a tool for performance experts, used in very specific analysis use-cases -- such as scheduling optimization~\cite{blass2021automatic} or message latency analysis~\cite{nishimura2021raplet} -- that were difficult to extend.
To improve the debugging experience in general, a more versatile approach is required.
Second, we need to ensure that the performance requirements of robotics are met.
Standard \ROStwo targets \emph{soft} real-time systems, i.e., systems where reaction time should have an upper bound, and should be achieved almost always, even though exceeding the bound is not catastrophic.
To maintain these characteristics, a tracing integration must have comparatively low overhead with very few outliers.

\ifx\outputstyle\arxiv
    \IEEEpubidadjcol
\fi

\textbf{Contributions.}
In this paper, we present \rostwotracing, a framework for tracing \ROStwo~\ifx\outputstyle\ieee\cite{ros2docs}\else\cite{macenski2022ros2}\fi\xspace with a collection of multipurpose low-overhead instrumentation and flexible tracing tools.
This new tool enables a wider range of precise analyses that help understand an application execution.
The source code is available at: \repourl.
\rostwotracing brings the following contributions:
\begin{itemize}
\item It offers extensible tracing instrumentation capable of providing execution information on multiple facets of \ROStwo.
\item With a strategic two-phase instrumentation design and using a low-overhead tracer, it has a lower runtime overhead than current solutions, making it suitable for the real-time applications targeted by \ROStwo.
\item It enables more precise analyses using combined \ROStwo userspace and kernel space data as a whole.
\end{itemize}
Furthermore, notable \rostwotracing features include a close integration into the expansive suite of \ROStwo orchestration \& general usability tools, and an easily swappable tracer backend to support different operating systems or to switch to a tracer with other desired features.

This paper is structured as follows:
We survey related work in \cref{sec:related-work} and summarize relevant background information in \cref{sec:background}.
We then present our solution in \cref{sec:ros2-tracing} and discuss its analysis potential in \cref{sec:analysis}.
Thereafter, \cref{sec:evaluation} presents an evaluation of the runtime overhead of our solution.
Future work is outlined in \cref{sec:future-work}.
Finally, we conclude in \cref{sec:conclusion}.

\section{Related Work}\label{sec:related-work}

Tracing is an established approach for performance analysis, popular for operating system-level performance analysis and for distributed systems.
Its popularity is both due to the low overhead when not in use, which is often zero, and due to the extensive tool support.
An excellent overview, covering both cloud and operating system use-cases, is~\cite{gregg2020systems,gregg2022performance}.
However, while powerful, these tools arguably operate at an abstraction level that is too low to be practical for the average roboticist.

In the context of robotics and \ROSone, the earliest reported use of tracing was motivated by non-deterministic behavior of obstacle avoidance in a mobile robot, by one of the present authors~\cite{Luetkebohle2017}.
Using a model of the \ROSone navigation stack, and tracing based on LTTng~\cite{desnoyers2006lttng}, a lack of synchronization between the sensory data processing and motion control pathways could be identified.
This is a primary example of how non-deterministic effects can be hard to diagnose otherwise, since the magnitude of the effect was dependent on how the OS scheduled the threads involved, which also varied over the run-time of the system.
There were some attempts at generalizing this kind of tracing tooling in~\cite{ros1tracetools}, and deriving a message flow analysis~\cite{bedard2019messageflow}.
However, they were discontinued due to the emergence of \ROStwo and its potential for real-time applications~\cite{ros2designwhy,ros2designrealtime}.

Previous work has identified various open problems in \ROStwo.
Kronauer et al.~\cite{kronauer2021latency} investigated the end-to-end latency of communications and found that its overhead is up to 50\% compared to directly using DDS, the underlying middleware.
Similarly, Jiang et al.~\cite{jiang2020message} found that the message conversion cost is highly dependent on the complexity of the message structure.
Casini et al.~\cite{casini2019response} proposed a scheduling model that aims to bound the end-to-end latency of processing chains.
Furthermore, several previous contributions propose tools that use tracing to measure and/or improve message transmission latency, due to its importance for realizing low-latency distributed systems.
This includes the RAPLET tool by Nishimura et al.~\cite{nishimura2021raplet} for \ROSone, ROS-FM by Rivera et al.~\cite{rivera2020ros}, and ROS-Llama by Blass et al.~\cite{blass2021automatic}, both targeting \ROStwo.
The last two are monitoring tools, meaning that they use instrumentation to extract execution information, process it, and provide the results to users or act on them during runtime.
However, none of these tools support any use-cases beyond latency, and they all show significant overheads~(cf.~\cref{tab:summary-comparison}), which is due for some of them to the use of logs or custom unoptimized tracers to extract execution information.
An interesting custom proposal to detect latency deadline violations with a low overhead of only 86~\micro\second{} per event, including online detection, is proposed by Peeck et al.~\cite{peeck2021online}, but again, without attempt at generality.

In conclusion, the present work is -- to the best of our knowledge -- the first and only tool that provides a generic, tracing-based approach for \ROStwo performance analysis.

\section{Background}\label{sec:background}

In this section, we summarize relevant background information needed to support subsequent sections.
Note that we use ``ROS~1'' to refer to the first version of ROS and use ``ROS'' to refer to \ROSone and \ROStwo in general, since many concepts apply to both.

\subsection{ROS 2 Architecture}\label{subsec:ros-2-architecture}

The \ROStwo architecture has multiple abstraction layers; from top to bottom, or user code to operating system: \texttt{rclcpp}/\texttt{rclpy}, \texttt{rcl}, and \texttt{rmw}.
The user code is above and the middleware implementation is below.
\texttt{rclcpp} and \texttt{rclpy} are the C++ and Python client libraries, respectively.
They are supported by \texttt{rcl}, a common client library API written in C which provides basic functionality.
The client libraries then implement the remaining features needed to provide the application-level API.
This includes implementing executors, which are high-level schedulers used to manage the invocation of callbacks (e.g., timer, subscription, and service) using one or more threads.
\texttt{rmw} is the middleware abstraction interface.
Each middleware that is used with \ROStwo has an implementation of this interface.
Multiple middleware implementations can be installed on a system, and the desired implementation can be selected at runtime through an environment variable, otherwise the default implementation is used.
As of \ROStwo Galactic Geochelone, the default middleware is Eclipse Cyclone DDS~\cite{eclipsecyclone}.

\subsection{ROS Nodes and Packages}\label{subsec:ros-nodes-packages}

ROS is based on the publish-subscribe paradigm and also supports the RPC pattern under the ``service'' name.
ROS nodes may both publish typed messages on topics and subscribe to topics, and they can use and provide services.
While the granularity and semantics of nodes in a system is a design choice, the resulting node and topics structure is analogous to a computation graph.

There are also specialized nodes called ``lifecycle nodes''~\cite{ros2designmanaged}.
They are stateful managed nodes based on a standard state machine.
This makes their life cycle easier to control, which can be beneficial for safety-critical applications~\cite{ros2designrealtime}.
Node life cycles can also be split into initialization phases and runtime phases, with dynamic memory allocations and other non-deterministic actions being constrained to the initialization phases for real-time applications.

As for the code, in ROS, it is generally split into multiple packages, which directly and indirectly depend on other packages.
Each package has a specific purpose and may provide multiple libraries and executables, with each executable containing any number of nodes.
The ROS ecosystem is federated: packages are spread across multiple code repositories, on various hosts, and are authored and maintained by different people.
The \ROStwo source code itself is made up of multiple packages that approximately match its architecture.

\subsection{Usability and Orchestration Tools}\label{subsec:orchestration-tools}

Much like \ROSone, \ROStwo has many tools for introspection, orchestration, and general usability.
There are various \texttt{ros2} commands, including \texttt{ros2 run} to run an executable and \texttt{ros2 topic} to manually publish messages and introspect published messages.
Packages can also provide extensions that add other custom commands.
The \texttt{ros2 launch} command is the main entry point for the \ROStwo orchestration system and allows launching multiple nodes at once.
This is configured through Python, XML, or YAML launch files which describe the system to be launched using nodes from any package or even other launch files.
Since ROS systems can be quite complex and contain multiple nodes, an orchestration system is indispensable.
Launch files can also be used to orchestrate test systems and verify certain behaviors or results.

\subsection{Generalizability}\label{subsec:generalizability}

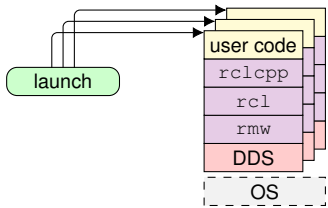
\begin{figure}[htbp]
\begin{center}
\tikzset{>={Latex[width=1.5mm,length=1.5mm]},
  baseTxt/.style = {scale=0.75, text centered, font=\sffamily},
  rosCoreTxt/.style = {baseTxt, font=\ttfamily},
  base/.style = {baseTxt, rectangle, draw=black, minimum width=2.0cm, minimum height=1cm},
  os/.style = {base, fill=gray!10, dashed},
  dupl/.style = {base, double copy shadow={shadow xshift=0.15cm, shadow yshift=0.15cm}},
  dds/.style = {dupl, fill=red!20},
  rosCore/.style = {dupl, fill=violet!20},
  userPkg/.style = {dupl, fill=yellow!20},
  launch/.style = {base, rounded corners, fill=green!20},
  launchArrows/.style = {rounded corners=0.75mm}
}

\begin{tikzpicture}[align=center]
  \draw[os]      (0,0.00)    rectangle (2.15,0.50)    node[baseTxt, pos=.5]    {OS};
  \draw[dds]     (0,0.60)  rectangle (1.75,1.10) node[baseTxt, pos=.5]    {DDS};
  \draw[rosCore] (0,1.10) rectangle (1.75,1.60)  node[rosCoreTxt, pos=.5] {rmw};
  \draw[rosCore] (0,1.60)  rectangle (1.75,2.10) node[rosCoreTxt, pos=.5] {rcl};
  \draw[rosCore] (0,2.10) rectangle (1.75,2.60)  node[rosCoreTxt, pos=.5] {rclcpp};
  \draw[userPkg] (0,2.60)  rectangle (1.75,3.10) node[baseTxt, pos=.5]    {user code};
  
  \draw[launch] (-1.5,1.95) rectangle (-3.5,2.45) node[baseTxt, pos=.5] (launch) {launch};
  
  \draw[->, launchArrows] ($(launch.north)-(0.15,0.2)$) |- (0,2.30);
  \draw[->, launchArrows] ($(launch.north)-(0.00,0.2)$) |- (0.15,2.45);
  \draw[->, launchArrows] ($(launch.north)+(0.15,-0.2)$) |- (0.3,2.60);
\end{tikzpicture}
 \end{center}
\vspacefiguretikz
\caption{Overall \ROStwo architecture and tooling interaction.}
\label{fig:background-overall-architecture}
\end{figure}

\cref{fig:background-overall-architecture} shows a summary of the \ROStwo architecture and the main tooling interaction.
The architecture and launch system can be generalized down to an orchestration tool managing an application layer on top of a middleware.
Therefore, the tool presented in this paper could be applied to other similar robotic systems.

\section{\rostwotracing}\label{sec:ros2-tracing}

In this section, we present the design and content of \rostwotracing.
It contains multiple ROS packages to support three different but complementary functionalities: instrumentation, usability tools, and test utilities.
\cref{tab:summary-comparison} shows a comparison between our proposed method and the existing methods mentioned in \cref{sec:related-work}.

\begin{table*}[htbp]
\begin{center}
\caption{Summary of Existing Monitoring and Instrumentation Methods and Comparison with Proposed Method}
\begin{minipage}{\textwidth}
\begin{tabular}{p{2.23cm}ccccccccccc}
\toprule
\multirow{2}{*}{\textbf{Method}}                        & \multirow{2}{*}{\textbf{Type\footnote{\footnotetextspace M and I for monitoring and instrumentation-only types, respectively.}}} & \multirow{2}{*}{\textbf{\begin{tabular}[c]{@{}c@{}}ROS\\ 1 / 2\end{tabular}}} & \multicolumn{5}{c}{\textbf{Instrumentation}}                         & \multirow{2}{*}{\textbf{Extensible}} & \multirow{2}{*}{\textbf{\begin{tabular}[c]{@{}c@{}}Launch\\ tools\end{tabular}}} & \multirow{2}{*}{\textbf{Overhead\footnote{\footnotetextspace C and L for CPU and latency overhead, respectively.}}} & \multirow{2}{*}{\textbf{\begin{tabular}[c]{@{}c@{}}Source\\ avail.\end{tabular}}} \\
                                                        &                                &                                                                               & {\footnotesize messages}   & {\footnotesize callbacks} & {\footnotesize services} & {\footnotesize executor} & {\footnotesize lifecycle}                               &                                      &                                                                                  &                                  &                                                                                      \\
\midrule
ROS-FM~\cite{rivera2020ros}                             & M                              & 1 / 2                                                                         & \checkmark                 & \texttimes                & \checkmark               & - / \texttimes           & - / \texttimes                                          & \texttimes                           & \texttimes                                                                       & 15-515\% C                       & \texttimes                                                                           \\
tracetools~\cite{ros1tracetools, bedard2019messageflow} & I                              & 1                                                                             & \checkmark                 & \checkmark                & \texttimes               & -                        & -                                                       & \texttimes                           & \texttimes                                                                       & ?                                & \checkmark                                                                           \\
RAPLET~\cite{nishimura2021raplet}                       & I                              & 1                                                                             & \checkmark                 & \checkmark                & \texttimes               & -                        & -                                                       & \texttimes                           & \texttimes                                                                       & 2-20\% L                         & \checkmark                                                                           \\
ROS-Llama~\cite{blass2021automatic}                     & M                              & 2                                                                             & \checkmark                 & \checkmark                & \texttimes               & \checkmark               & \texttimes                                              & \texttimes                           & \texttimes                                                                       & 30-40\% C                        & \texttimes                                                                           \\
\rostwotracing                                          & I                              & 2                                                                             & \checkmark                 & \checkmark                & \checkmark               & \checkmark               & \checkmark                                              & \checkmark                           & \checkmark                                                                       & 1-15\% L                         & \checkmark                                                                           \\
\bottomrule
\end{tabular}
\label{tab:summary-comparison}
\vspacetablefootnotes
\end{minipage}
\end{center}
\end{table*}

\subsection{Instrumentation}\label{sec:ros2-tracing-instrumentation}

\begin{figure}[htbp]
\begin{center}
\tikzset{>={Latex[width=1.5mm,length=1.5mm]},
  baseTxt/.style = {scale=0.75, text centered, font=\sffamily},
  base/.style = {baseTxt, rectangle, rounded corners, draw=black, minimum width=1.5cm, minimum height=0.75cm},
  pkg/.style = {base, fill=violet!20},
  tracetools/.style = {base, fill=yellow!20, font=\ttfamily},
  tracer/.style = {base, fill=red!20, dashed}
}

\begin{tikzpicture}[node distance=5.0cm, align=center]
  \node (instrumentedPkg) [pkg] {ROS 2 \\ package};
  \node (tracetools) [tracetools, right of=instrumentedPkg] {tracetools};
  \node (tracerMain) [tracer, right of=tracetools] {Tracer};
  \node (tracerSecond) [tracer, below of=tracerMain, yshift=4.00cm] {Tracer};

  \draw[->] (instrumentedPkg) -- node [baseTxt, above] {instrumentation} node [baseTxt, below] {function call} (tracetools);
  \draw[->, dashed] (tracetools) -- node [baseTxt, above] {tracepoint} node [baseTxt, below] {call} (tracerMain);
  \draw[->, dashed] ($(tracetools.east)+(0.3,0)$) |- (tracerSecond);
\end{tikzpicture}
 \end{center}
\vspacefiguretikz
\caption{Instrumentation and tracepoint calls.}
\label{fig:pkg-instrumentation}
\end{figure}
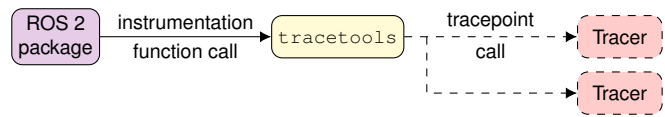

As shown in \cref{fig:pkg-instrumentation}, core \ROStwo packages are instrumented with function calls to the \texttt{tracetools} package.
This package provides tracepoints for all of \ROStwo and is the one that triggers them.
Tracepoints usually act as instrumentation points and could be directly added to the instrumented code.
This creates an indirection, which we introduced for two complementary reasons.
First, it allows for abstracting away the tracer backend and allows to easily switch the tracer.
Indeed, by replacing the compiled \texttt{tracetools} library, the tracer backend can be replaced without affecting the instrumented packages (i.e., the core \ROStwo code).
This could be done to support tracing on a different platform or to use a tracer that has other desired functionalities.
Second, this keeps instrumented packages free of boilerplate code (e.g., tracepoints definitions and other required preprocessor macros).
However, the main advantage of this design choice also has a slight downside, since adding new tracepoints requires modifying both the instrumented package and the \texttt{tracetools} package.

As shown in \cref{tab:summary-comparison}, in addition to being extensible, our method provides instrumentation for multiple aspects of \ROStwo, including messages, callbacks, services, executor states, and lifecycle states.
\cref{tab:tracepoints} presents a full list of the instrumentation points provided by \texttt{tracetools}.
We split the instrumentation points into two types: initialization events and runtime events.
The former collect one-time information about the state of objects, e.g., creation of publishers, subscriptions, and services.
The latter collect information about events throughout the runtime, e.g., message publication and callback execution.
The former are therefore predominantly triggered during the system initialization phase and are used to minimize the payload size of the latter.
This strategic instrumentation design is key to minimizing the overhead in the runtime phase.
For example, all publisher-related tracepoints in the runtime phase include a unique identifier for the publisher, which is then matched with the data collected by the publisher-related tracepoints in the initialization phase (e.g., topic name, corresponding node name, etc.), thus minimizing the payload size of runtime tracepoints.
The information that is collected to form the trace data can then be used to build a model of the execution.
Due to the very abstractional nature of the \ROStwo architecture, multiple instrumentation points are sometimes needed to gather the necessary information.
For example, to build a model of a subscription, we collect information about its callback function from \texttt{rclcpp} and its topic name from \texttt{rcl}.
This is because callbacks are managed by the client library.
Furthermore, both the instrumentation point name and the payload are meaningful: some instrumentation points only differ by their names and are used to indicate the originating layer.
Since the instrumentation points cover multiple analysis use-cases, if a portion of the information is not needed for a given analysis, some of the instrumentation points can be disabled and therefore have virtually no impact on execution.

We did not instrument the Python client library, since it is not used for the kind of real-time applications that we are considering with \rostwotracing.
However, we instrumented \texttt{rcl} directly whenever possible.
By putting the instrumentation as low as possible in the \ROStwo architecture, it can be leveraged to more easily support other client libraries in the future.

The \textit{Linux Trace Toolkit: next generation} (LTTng) tracer was chosen as the default tracing backend, for its low overhead and real-time compatibility as well as its ability to trace both the kernel and userspace~\cite{desnoyers2006lttng, fournier2009combined}.
The runtime cost per LTTng userspace tracepoint on a vanilla Linux kernel using an Intel i7-3770 CPU (3.40~GHz) with 16~GB of RAM is approximately 158~ns~\cite{gebai2018survey}.
Since it is a Linux-only tracer, all instrumentation calls to the \texttt{tracetools} package are preprocessed out on other platforms.
This can also be achieved on Linux through a build option.

\begin{table}[htbp]
\begin{center}
\caption{Instrumentation Points List with Types}
\begin{minipage}{\columnwidth}
\begin{tabular}{clcc}
\toprule
\textbf{ROS 2 Layer} & \textbf{Instrumentation Point Name} & \textbf{Type\footnote{\footnotetextspace I and R for initialization and runtime types, respectively.}} & \textbf{Note\footnote{\footnotetextspace P and S for publishing and message reception hot paths, respectively.}} \\
\midrule
\multirow{13}{*}{\texttt{rclcpp}} & rclcpp\_subscription\_init            & I & \\
                                  & rclcpp\_subscription\_callback\_added & I & \\
                                  & rclcpp\_publish                       & R & P \\
                                  & rclcpp\_take                          & R & S \\
                                  & rclcpp\_service\_callback\_added      & I & \\
                                  & rclcpp\_timer\_callback\_added        & I & \\
                                  & rclcpp\_timer\_link\_node             & I & \\
                                  & rclcpp\_callback\_register            & I & \\
                                  & callback\_start                       & R & S \\
                                  & callback\_end                         & R & \\
                                  & rclcpp\_executor\_get\_next\_ready    & R & S \\
                                  & rclcpp\_executor\_wait\_for\_work     & R & S \\
                                  & rclcpp\_executor\_execute             & R & S \\
\hline
\multirow{11}{*}{\texttt{rcl}} & rcl\_init                            & I & \\
                               & rcl\_node\_init                      & I & \\
                               & rcl\_publisher\_init                 & I & \\
                               & rcl\_subscription\_init              & I & \\
                               & rcl\_publish                         & R & P \\
                               & rcl\_take                            & R & S \\
                               & rcl\_client\_init                    & I & \\
                               & rcl\_service\_init                   & I & \\
                               & rcl\_timer\_init                     & I & \\
                               & rcl\_lifecycle\_state\_machine\_init & I & \\
                               & rcl\_lifecycle\_transition           & R & \\
\hline
\multirow{4}{*}{\texttt{rmw}} & rmw\_publisher\_init    & I & \\
                              & rmw\_subscription\_init & I & \\
                              & rmw\_publish            & R & P \\
                              & rmw\_take               & R & S \\
\bottomrule
\end{tabular}
\label{tab:tracepoints}
\vspacetablefootnotes
\end{minipage}
\end{center}
\end{table}

\subsection{Usability Tools}

In line with the \ROStwo usability and orchestration tools, our proposed solution includes two different interfaces to control tracing: a \texttt{ros2 trace} command and a \texttt{Trace} action for launch files.
The \texttt{ros2 trace} command is a simple command that allows configuring the tracer to start tracing.
The system or executable to be traced must then be run or launched in a separate terminal.
When the application is done running, tracing must be stopped in the original \texttt{ros2 trace} terminal.
On the other hand, the \texttt{Trace} action can be used in XML, YAML, and Python launch files.
It then allows configuring the tracer and launching the system at the same time.
Tracing is stopped automatically after the launched system has shut down, either on its own or after being manually terminated.
\cref{lst:trace-launch} shows an example with an XML file that launches two nodes.
While \ROStwo does not currently natively support it, this would be useful for remote or multi-host orchestration to trace all hosts at once and aggregate the resulting traces.

\begin{lstlisting}[
  caption={\texttt{Trace} action in XML launch file},
  label={lst:trace-launch},
  language=XML,
  basicstyle=\ttfamily\small,
  belowskip=2em,  %
  frame=single
]
<launch>
  <trace
    session-name="ros2" events-ust="ros2:*"/>
  <node pkg="package_a" exec="executable_x"/>
  <node pkg="package_b" exec="executable_y"/>
</launch>
\end{lstlisting}

Furthermore, these tools can be used to leverage existing LTTng instrumentation (e.g., kernel and other userspace instrumentation) and to enable any custom application-level tracepoints to record other relevant data.
For example, LTTng provides shared libraries that can be preloaded with \texttt{LD\_PRELOAD} to intercept calls to \texttt{libc}, \texttt{pthread}, the dynamic linker, and function entry \& exit instrumentation (added with \texttt{-finstrument-functions}) and trigger tracepoints before calling the real functions.
If those tracepoints are enabled through launch files, the corresponding shared libraries will be located and preloaded automatically for all executables, which greatly simplifies the launch configuration.
The tools also do not prevent users from directly configuring the tracer for advanced options.
They are only a thin flexible compatibility and usability layer for \ROStwo and use the LTTng Python bindings for tracer control.

\subsection{Test Utilities}

The \texttt{tracetools\_test} package provides a test utility that allows running nodes and tracing them.
The resulting trace can then be read in the test using the \texttt{tracetools\_read} package to assert results or behaviors in a lower-level, less invasive way.

\section{Analysis}\label{sec:analysis}

\begin{figure}[htbp]
\centerline{\includegraphics[width=\columnwidth]{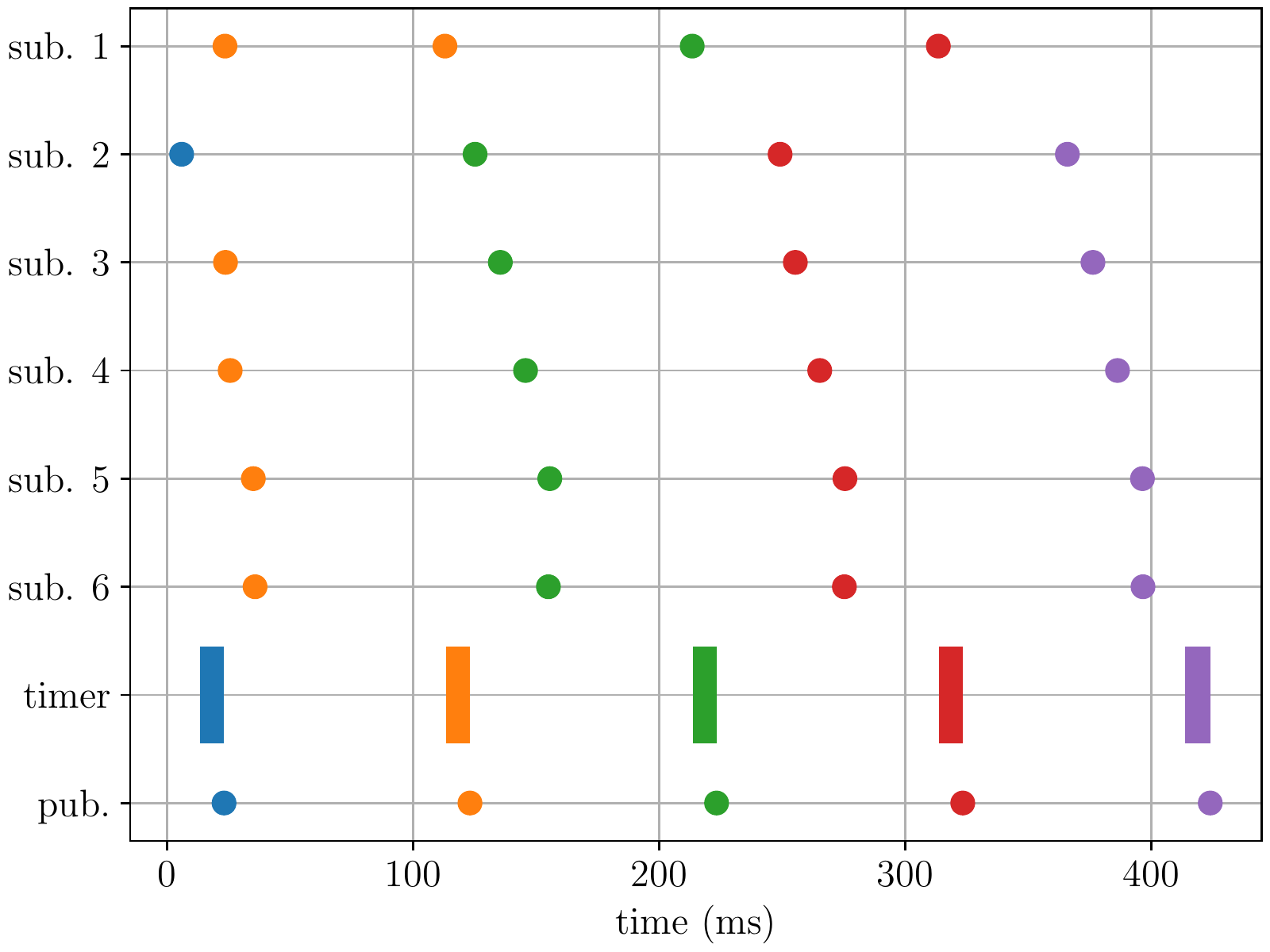}}
\vspacefigureimage
\caption{
Example time chart of subscription message reception (sub.), timer callback execution (timer), and message publication (pub.).
Message reception and publication instances are displayed as single timestamps, while timer callback executions are displayed as ranges, with a start and an end.
The periodic timer callback uses the last received message from each subscription to compute a result and publish it; this inputs-outputs link is illustrated using colors, highlighting an inadequate synchronization.
}
\label{fig:analysis-ros2-chart}
\end{figure}

\begin{figure}[htbp]
\centerline{\includegraphics[width=\columnwidth]{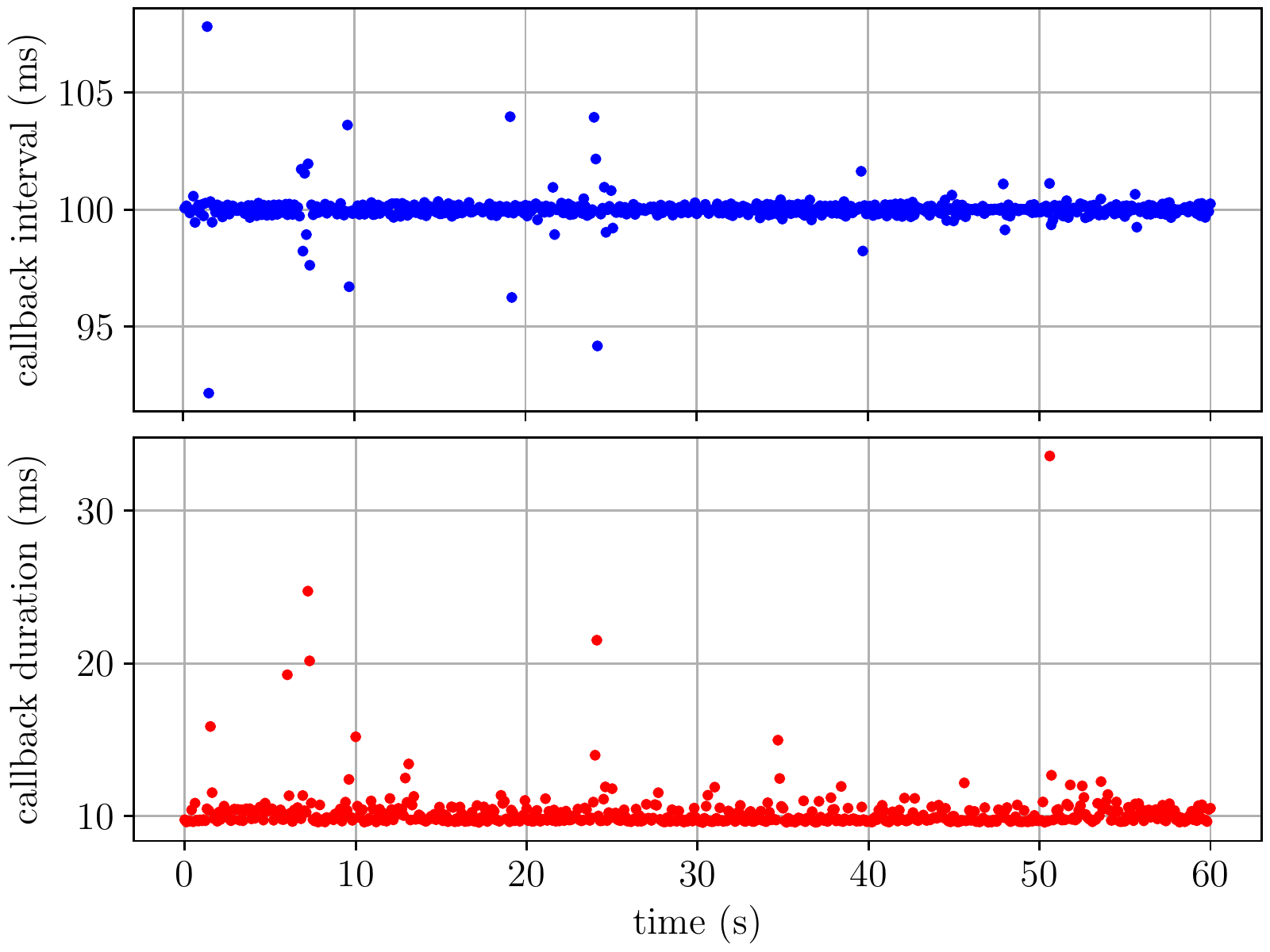}}
\vspacefigureimage
\caption{
Example timer callback execution interval~(top) and duration~(bottom) over time.
The callback period is set to 100~ms, while the callback duration depends on the work done.
Both contain outliers.
}
\label{fig:analysis-ros2-timer}
\end{figure}

The instrumentation and tracing tools provided by \rostwotracing allow collecting execution information at the \ROStwo level.
This information can then be processed using existing tools to compute metrics or to provide models of the execution.
For example, we traced a \ROStwo system that simulates a real-world autonomous driving scenario~\cite{rtwgreferencesystem}.
In this example, a node receives data from 6 different topics using subscriptions.
When the subscriptions receive a new message, they cache it so that the node only keeps the latest message for each topic.
The periodically-triggered callback uses the latest message from each topic to compute a result and publish it.
To analyze the trace data, we wrote a simple script using \tracetoolsanalysis~\cite{tracetoolsanalysis}, a Python library to read and analyze trace data.
As shown in \cref{fig:analysis-ros2-chart}, we can display message reception and publication instance timestamps as well as timer callback execution intervals over time.
There is a visible gradual time drift between the input and output messages, which could impact the validity of the published result, similar to the issue described by~\cite{Luetkebohle2017}.
This could warrant further analysis and tuning, depending on the system requirements.
We can also compute and display the timer callback execution interval and duration, as shown in \cref{fig:analysis-ros2-timer}.
The timer callback period is set to 100~ms and the duration is approximately 10~ms, but there are outliers.
This jitter could negatively affect the system; these anomalies could warrant further analysis as well.

\begin{figure*}[!t]
\centerline{\includegraphics[width=\textwidth]{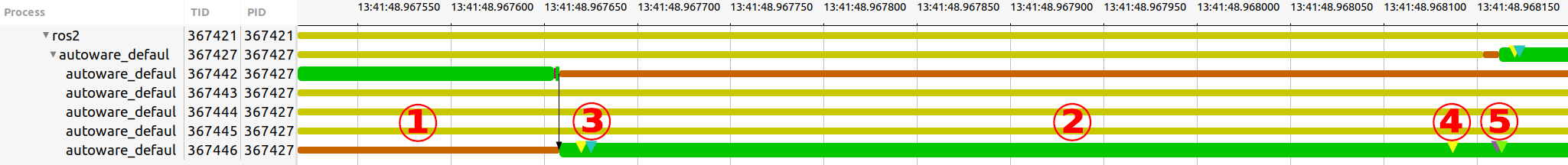}}
\vspacefigureimage
\caption{
State of \ROStwo application threads over time with timestamps of \ROStwo events from the \rostwotracing instrumentation displayed as small triangles on top of the thread state rectangle:
\circled{1} thread waiting for CPU for 9.9~ms (orange), \circled{2} thread running (green), \circled{3} event marking start of middleware query \& wait for new messages, \circled{4} event marking end of middleware query, and \circled{5} \texttt{rclcpp\_executor\_execute} event followed shortly after by \texttt{callback\_start} event for timer callback.
This result was obtained by importing trace data collected from the Linux kernel and from the \ROStwo application using LTTng into Trace Compass~\cite{tracecompass}.
The black arrow to the left of \circled{3} represents the scheduling switch from one thread to another for a given CPU.
Some less relevant threads were hidden.
}
\label{fig:analysis-tc-control-flow}
\end{figure*}

To dig deeper, this information can be paired with data from the operating system: OS trace data can help find the cause of performance bottlenecks and other issues~\cite{cote2016problem}.
Since \ROStwo does higher-level scheduling, this is critical for understanding the actual execution at the OS level.
Using LTTng, the Linux kernel can be traced alongside the application.
The \ROStwo trace data that was obtained using the \rostwotracing instrumentation can be analyzed together with the OS trace data using Eclipse Trace Compass~\cite{tracecompass}, which is an open-source trace viewer and framework aimed towards performance and reliability analysis.
Trace Compass can analyze Linux kernel data to show the frequency and state of CPUs over time, including interrupts, system calls, or userspace processes executing on each CPU.
It can also display the state of each thread over time, as shown in \cref{fig:analysis-tc-control-flow} for the application threads.
Building Trace Compass analyses specific to \ROStwo is left as future work; however, we can visualize timestamps of \ROStwo trace events on top of the existing analyses.
From the \ROStwo trace data shown in \cref{fig:analysis-ros2-timer}, we know that the timer callback instance following the longest interval is at the 1.4~s mark with 107.8~ms.
Finding the timestamps of the corresponding \ROStwo events in Trace Compass, we see that the thread was blocked waiting for CPU for 9.9~ms before the aforementioned callback instance.
The thread then had to query the middleware for new messages (even if timers are strictly handled at the \ROStwo level) and finally call the overdue timer callback.
In this example, a multi-threaded executor was used, with the number of threads being equal to the number of logical CPU cores by default.
Since this was not the only application running on the system at that time, multiple processes and threads were competing for CPU, as can be observed using Trace Compass.
Therefore, the executor settings could be tuned, or the executor could be replaced by another type of executor with features that better meet the requirements for this system, which could entail creating a new executor: this is an open problem in \ROStwo.
A multi-level analysis such as this one would not have been possible without collecting both userspace \& kernel execution information, and analyzing the combined data, which current tools do not offer.
The scripts and full instructions to replicate this example are available at: \href{https://github.com/christophebedard/ros2_tracing-analysis-example}{github.com/christophebedard/ros2\_tracing-analysis-example}.

\section{Evaluation}\label{sec:evaluation}

To validate that our proposed solution is compatible latency-wise with real-time systems, we evaluate the overhead of \rostwotracing, or specifically its instrumentation overhead.
This is the time consumed by the instrumentation within the monitored process.
When enabled, it directly affects these processes by adding latency.

Since the \rostwotracing tracepoints are placed along the message publication and reception pipeline, an easy way to capture the maximum overhead is to measure the time between publishing a message and when it is handled by the subscription callback.
This is what we will do in the following.

\subsection{Experiment Setup}\label{subsec:experiment-setup}

We use the standard message-passing latency benchmark for \ROStwo, \performancetest~\cite{performancetest}, with a minimal configuration: one publisher node and one subscription node.
We vary message size and publishing rate, since it is known that they affect middleware performance~\cite{kronauer2021latency}.
The parameter space is shown in \cref{tab:experiment-parameters} and is based on typical use-cases~\cite{reke2020self}.

\begin{table}[htbp]
\begin{center}
\caption{Experiment Parameters and Values}
\begin{tabular}{cc}
\toprule
\textbf{Publishing rate (Hz)} & 100, 500, 1000, 2000 \\
\hline
\textbf{Message size (KiB)} & 1, 32, 64, 256 \\
\hline
\textbf{Quality of service} & reliable only \\
\hline
\textbf{DDS implementation} & eProsima Fast DDS \\
\bottomrule
\end{tabular}
\label{tab:experiment-parameters}
\end{center}
\end{table}

To reduce measurement variability, we follow common practice by using a kernel patched with the PREEMPT\_RT patch (5.4.3-rt1), disabling simultaneous multithreading, and disabling power-saving features in the BIOS (dynamic frequency scaling, C-states, Turbo Boost, etc.).
Further, we increase UDP buffers to 64~MB to ensure sufficient networking performance for larger messages.
The experiment was run on an Intel i7-3770 (3.40~GHz) 4-core CPU, 8~GB RAM system with Ubuntu 20.04.2.
All measurements are based on the \ROStwo Rolling distribution, in between the Galactic and Humble releases, which is the most recent version at this time.
While Eclipse Cyclone DDS~\cite{eclipsecyclone} is the default DDS implementation for the current \ROStwo release, Galactic, we have found eProsima Fast DDS~\cite{eprosimafastdds} to be more stable for larger messages, and it is also the default for the upcoming release, Humble.
We have therefore used Fast DDS.

For each combination in the parameter space, \performancetest is run for 60 minutes and scheduled with the SCHED\_FIDO real-time policy with the highest priority (99).
To reduce outliers due to system initialization, the first 10 seconds of each recording are discarded.
To determine the tracing overhead, we run the experiment once without any tracing enabled, and once with all tracepoints enabled.
In practice, since not all tracepoints might be needed depending on the intended analysis, this represents the worst-case scenario.

The code and instructions to replicate this experiment are available at: \href{https://github.com/christophebedard/ros2_tracing-overhead-evaluation}{github.com/christophebedard/ros2\_tracing-overhead-evaluation}.

\subsection{Results and Discussion}\label{subsec:results-discussion}

\begin{figure}[htbp]
\centerline{\includegraphics[width=\columnwidth]{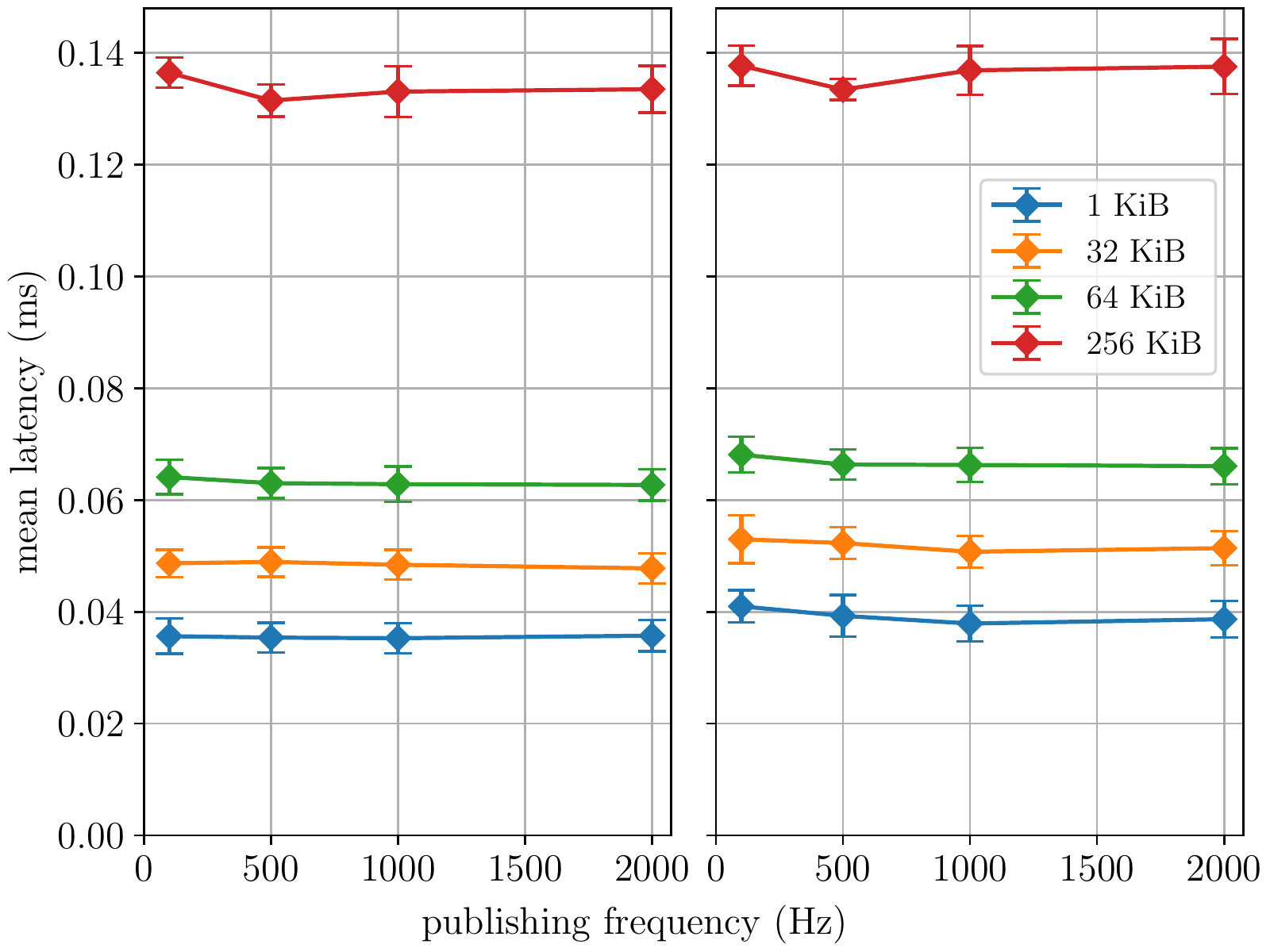}}
\vspacefigureimage
\caption{Message latencies (avg. $\pm$ std.) without tracing~(left) and with tracing~(right).}
\label{fig:experiment-latencies}
\end{figure}

\begin{figure}[!htbp]
\centerline{\includegraphics[width=\columnwidth]{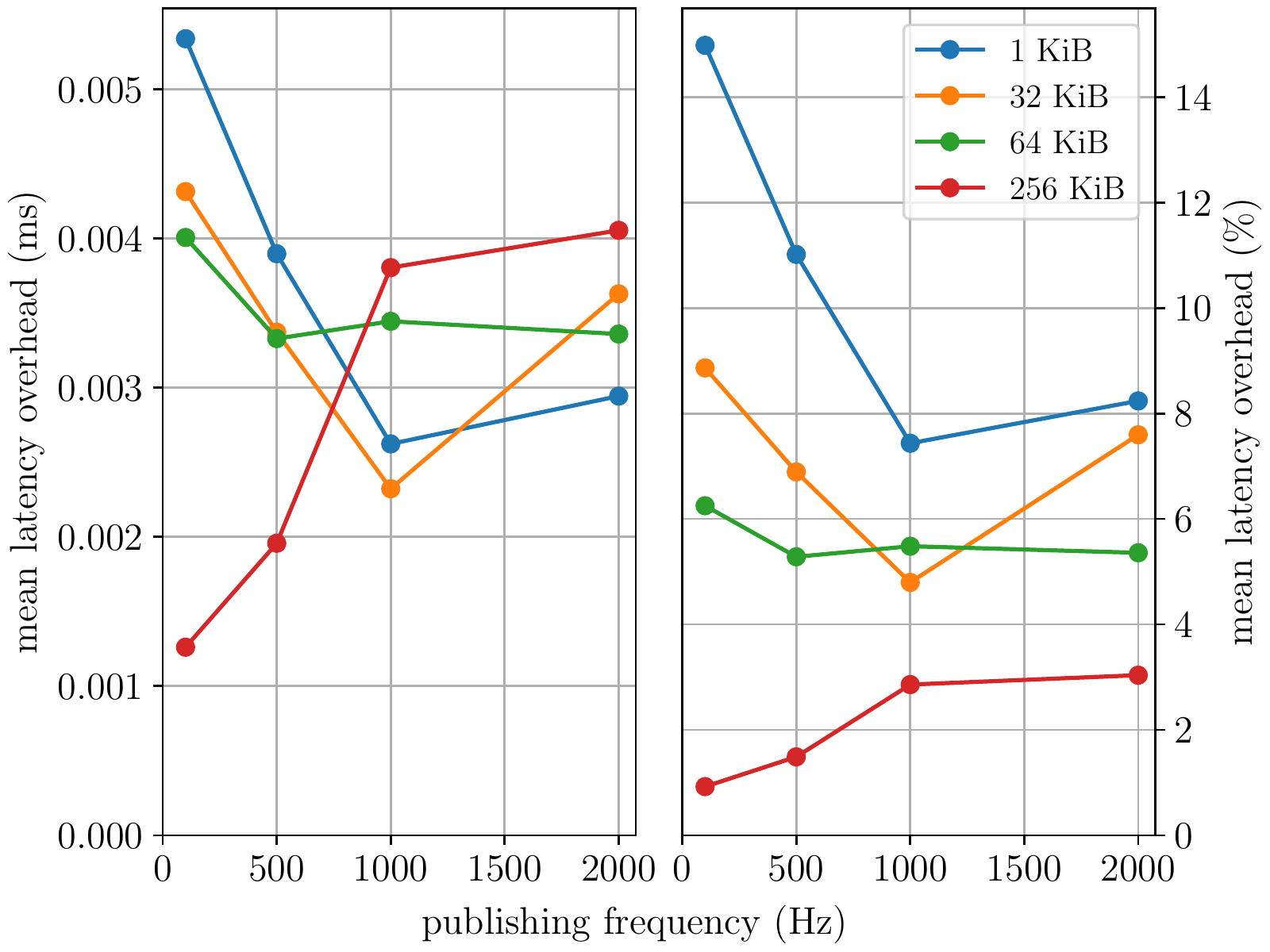}}
\vspacefigureimage
\caption{
Absolute~(left) and relative~(right) latency overhead results.
The standard deviation of the difference between the two means is insignificant here.
}
\label{fig:experiment-overhead}
\end{figure}

\cref{fig:experiment-latencies} shows the individual average latencies without and with tracing, while \cref{fig:experiment-overhead} shows the absolute and relative latency overhead.

First, as expected, the mean latency values increase with the message size, and the relative latency overhead values decrease with the message size.
There is no significant decrease in latency as the publishing frequency increases; this behavior was however much more noticeable before disabling power-saving features through the BIOS.
We would also expect the CPU overhead to be the same for all message sizes and publishing frequencies, since it is, in theory, a constant overhead for message publication and reception.
However, it can be seen that this is not the case, and instead, the absolute overhead is larger at small frequencies.
This is somewhat puzzling, and it would certainly merit further experiments.
However, due to the overall small effect, we are approaching a range where cache effects in the CPU or other non-deterministic factors come into play.
Since the overhead values are overall fairly close, the overhead does not seem to be related to any of the experiment parameters, and the absolute values are well within acceptable ranges, we consider the requirements set out for \rostwotracing to be fulfilled.
Additionally, these absolute latency overhead results are within one order of magnitude of the results that~\cite{gebai2018survey} presented: since there are 10 tracepoints in the publish-subscribe hot path (see \cref{tab:tracepoints}), the overhead should therefore be 10 $\cdot$ 158~ns $=$ 0.00158~ms, which is indeed comparable.

\begin{figure}[htbp]
\centerline{\includegraphics[width=\columnwidth]{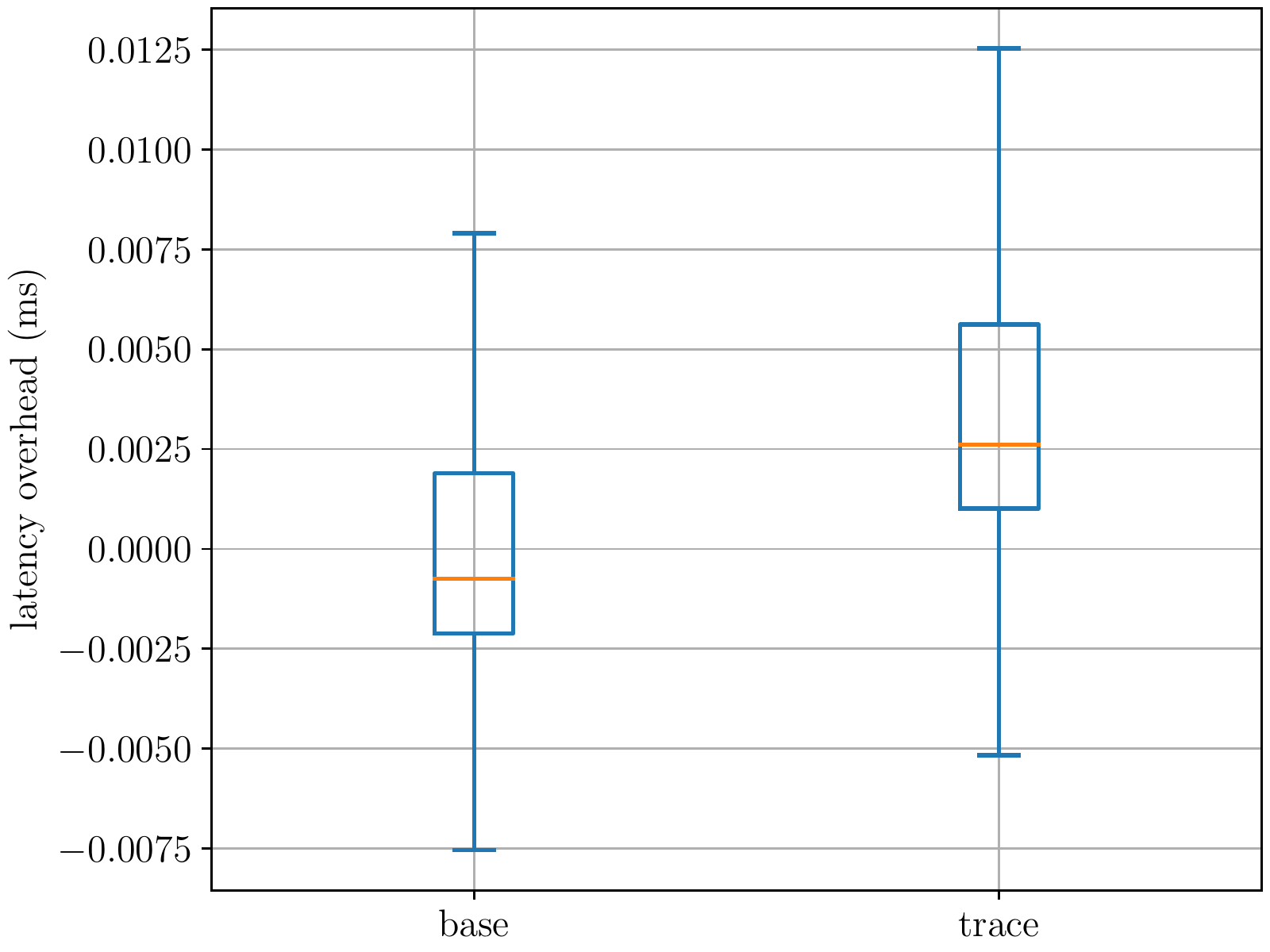}}
\vspacefigureimage
\caption{
Aggregated latency overhead and variation without tracing~(left) and with tracing~(right).
Latency values have been individually normalized to zero mean based on the latencies without tracing, showing overhead and variation.
Note that the left mean is very slightly below zero due to the additional imbalance caused by the variation in publishing frequency.
}
\label{fig:aggregate-latency}
\end{figure}

Since most practical systems use a mixture of message sizes and frequencies, we also analyze the overhead by aggregating it over all experiment runs.
For each combination of message size and publishing frequency, we use the two sets of latencies (i.e., without tracing and with tracing, represented in \cref{fig:experiment-latencies}), and subtract the mean of the no-tracing set from all latencies.
By aggregating the latency differences for all combinations, we obtain two sets of latency overheads, which are represented in \cref{fig:aggregate-latency} (without tracing and with tracing, respectively).
Note that the aggregate overhead is more strongly influenced by the higher publication frequencies, since more messages are sent in the same time frame.
The mean overhead is thus 0.0033~ms, with 50\% of the data between 0.0010~ms and 0.0056~ms.

No measurement system can be entirely without overhead~-- however, we think that these results show that \rostwotracing has very low overhead, which is acceptable for most target situations for \ROStwo.
It is certainly lower than known alternatives as presented in \cref{tab:summary-comparison}.

Nonetheless, for very busy systems or very particularly CPU-constrained platforms, overheads might add up enough to impact the messaging performance.
For such situations, there are two potential optimization options:
First, about half of the tracepoints would be sufficient for basic information, and, second, tracing can be selectively enabled on only some of the processes, instead of all \ROStwo nodes.
\section{Future Work}\label{sec:future-work}

Many improvements and additions could be made to \rostwotracing.
While the RPC pattern is not used as much in real-time applications, instrumentation could be added to support services and actions, with the latter being services with optional progress feedback.
Furthermore, \rostwotracing does not gather information about object destruction, e.g., if a publisher or a subscription is destroyed during runtime.
This is because it does not fit with design guidelines of real-time safety-critical systems, where the system is usually static once it enters its runtime phase.
Nonetheless, complete object lifetime information could be gathered.
Middleware implementations could also be instrumented to provide lower-level information on the handling of messages.
Instrumenting \texttt{rclc}~\cite{staschulat2020rclc, staschulat2021budget} would also be interesting.
It is a client library written in C with a deterministic executor aimed at \ROStwo applications on memory-limited real-time platforms such as microcontrollers.

As for the usability tools, as mentioned previously, the orchestration tools could be improved, when native support for remote or coordinated multi-host orchestration gets added to the \ROStwo launch system.

While the overall approach taken for \rostwotracing is primarily aimed at offline monitoring and analysis, the instrumentation itself could be leveraged for online monitoring.
For example, the LTTng live mode could be used to do online processing of the trace data.
Another backend could also be added for a tracer that better supports online monitoring.
There could also be other default backends for other operating systems, like QNX, which is often used for real-time as well.

In future work, we plan on building on the \rostwotracing instrumentation and tools to analyze the internal workings of \ROStwo.
For example, as mentioned in \cref{sec:related-work,,sec:analysis}, the determinism and efficiency of the \ROStwo executors could be analyzed and compared to proposed alternatives.
The \ROStwo instrumentation could of course also be used in conjunction with the LTTng built-in userspace and kernel instrumentation, as demonstrated in \cref{sec:analysis}.
For example, to verify real-time systems, unwanted runtime phase dynamic memory allocations could be detected by combining lifecycle node state information and the LTTng \texttt{libc} memory allocation tracepoints.
Trace Compass could also be used to provide analyses and views specific to \ROStwo.

\section{Conclusion}\label{sec:conclusion}

Testing and debugging robotic systems, based on recordings of high-level messages, does not provide sufficient information on the computation performed, to identify causes of performance bottlenecks or other issues.
Existing methods target very specific problems and thus cannot be used for multipurpose analysis.
They are also not suitable for real-world real-time applications, because of their high overhead or poor usability.

We presented \rostwotracing, a framework with instrumentation and flexible tools to trace \ROStwo.
The extensible multipurpose low-overhead instrumentation for the \ROStwo core allows collecting execution information to analyze any \ROStwo system.
The tools promote usability through their integration with the \ROStwo orchestration system and other usability tools.
Our experiments showed that the message latency overhead it introduces is in an acceptable range for real-time systems built on \ROStwo.
These tools enable testing and debugging \ROStwo applications based on internal execution information, in a manner that is compatible with real-time applications and real-world development processes.
Analyzing the combined trace data, from a \ROStwo application and the operating system, can help find the cause of performance bottlenecks and other issues.
We plan on leveraging \rostwotracing in future work to analyze the internal handling of \ROStwo messages.
 
\bibliographystyle{IEEEtran}

\end{document}